\documentclass[conference]{IEEEtran}
\IEEEoverridecommandlockouts
\usepackage{cite}
\usepackage{amsmath,amssymb,amsfonts}
\usepackage{algorithmic}
\usepackage{graphicx}
\usepackage{textcomp}
\usepackage{caption}
\usepackage{booktabs}
\usepackage{hyperref}
\usepackage{multirow}
\usepackage{caption}
\usepackage{makecell}
\usepackage[normalem]{ulem}
\useunder{\uline}{\ul}{}
\usepackage[table]{xcolor}
\def\BibTeX{{\rm B\kern-.05em{\sc i\kern-.025em b}\kern-.08em
    T\kern-.1667em\lower.7ex\hbox{E}\kern-.125emX}}

\hypersetup{
    colorlinks=true,    
    urlcolor=magenta,    
}

\begin{document}

\setlength{\abovedisplayskip}{3pt}
\setlength{\belowdisplayskip}{3pt}

\title{TextureDiffusion: Target Prompt Disentangled Editing for Various Texture Transfer}

\author{\IEEEauthorblockN{Zihan Su}
\IEEEauthorblockA{\textit{Shenzhen International Graduate School} \\
\textit{Tsinghua University}\\
Shenzhen, China \\
zh-su24@mails.tsinghua.edu.cn}
\and
\IEEEauthorblockN{Junhao Zhuang}
\IEEEauthorblockA{\textit{Shenzhen International Graduate School} \\
\textit{Tsinghua University}\\
Shenzhen, China \\
zhuangjh23@mails.tsinghua.edu.cn}
\and
\IEEEauthorblockN{Chun Yuan$^\dag$\thanks{$^\dag$ Corresponding author.}}
\IEEEauthorblockA{\textit{Shenzhen International Graduate School} \\
\textit{Tsinghua University}\\
Shenzhen, China \\
yuanc@sz.tsinghua.edu.cn}
}

\maketitle

\begin{abstract}
Recently, text-guided image editing has achieved significant success. However, existing methods can only apply simple textures like wood or gold when changing the texture of an object. Complex textures such as cloud or fire pose a challenge.
This limitation stems from that the target prompt needs to contain both the input image content and \emph{$<$texture$>$}, restricting the texture representation. 
In this paper, we propose \textit{TextureDiffusion}, a tuning-free image editing method applied to various texture
transfer. Initially, the target prompt is directly set to \emph{``$<$texture$>$"}, making the texture disentangled from the input image content to enhance texture representation. Subsequently, query features in self-attention and features in residual blocks are utilized to preserve the structure of the input image. Finally, to maintain the background, we introduce an edit
localization technique which blends the self-attention results and the intermediate latents.
Comprehensive experiments demonstrate that \textit{TextureDiffusion} can harmoniously transfer various textures with excellent structure and background preservation. Code is publicly available at \textit{\url{https://github.com/THU-CVML/TextureDiffusion}}
\end{abstract}


\section{Introduction}
Despite the powerful content generation capabilities of text-to-image generative models~\cite{ramesh2021zero,nichol2021glide,yu2022parti,ramesh2022hierarchical,balaji2022ediff,chen2022distribution}, there are still some limitations on the user's control over the generated images. In order to increase user's control, text-guided image editing is particularly important.

Existing text-guided image editing methods~\cite{yang2023dynamic,Imagic,Instructpix2pix,Text2live,diffusionclip,multiedits,p2p,pnp,infedit} can accomplish various editing tasks, such as object addition and removal, action change, and texture change. 
Prompt-to-Prompt (P2P)~\cite{p2p} found that the cross-attention map corresponded to the mapping relationship between text and image. Plug-and-Play (PnP)~\cite{pnp} injected the self-attention maps and features into the generation process of the target image to maintain the consistency of the spatial layout. 
InfEdit~\cite{infedit} introduced a virtual inversion strategy and unified attention control to facilitate consistent and accurate editing.

\begin{figure}[!t]
    \centering
    \includegraphics[width=\linewidth]{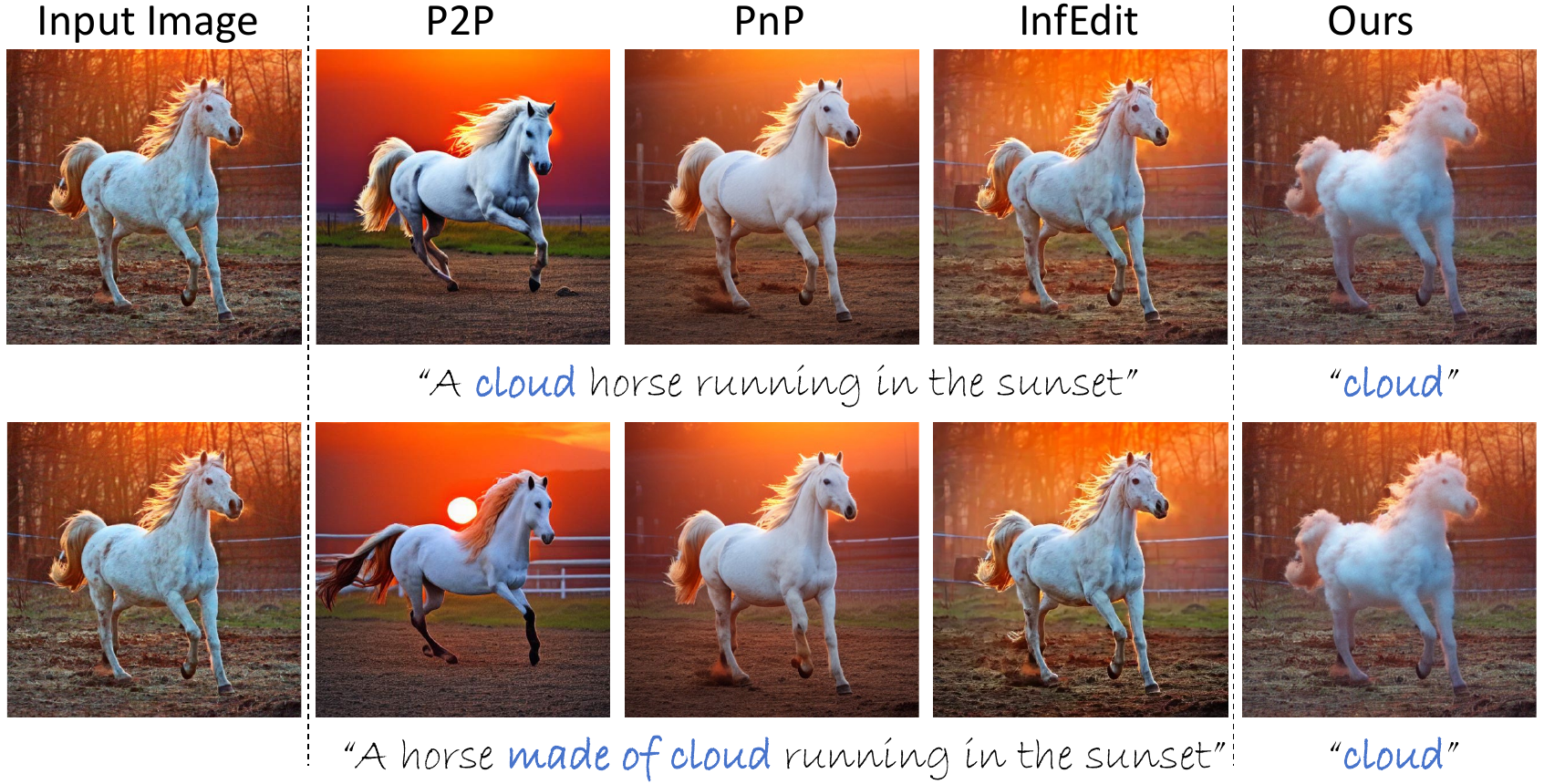}
    \caption{Existing text-guided image editing methods cannot transfer complex textures. By making the texture disentangled from the description of the input image in the target prompt and applying the proposed structure preservation module and edit localization technique, \textit{TextureDiffusion} can harmoniously transfer various textures to the target object.}
    \vspace{-0.5cm}
    \label{fig:teaser figure}
\end{figure}

However, for the texture transfer task, i.e., changing the texture of the target object, the previous methods are limited to simple textures like wood or gold. 
The challenge arises when attempting to transfer more complex textures, such as cloud or fire.
When describing \textit{\textless texture\textgreater} \ in the target prompt, ``wood" corresponds to ``wooden" and ``gold" corresponds to ``golden", but there is no corresponding adjective for ``cloud". If ``cloud" is forced to be included in the text description, the previous methods cannot successfully transfer the texture, as shown in Fig.~\ref{fig:teaser figure}.
This limitation stems from that the target prompt needs to contain both the input image content and \textit{\textless texture\textgreater}, restricting the texture representation.

\begin{figure*}[!t]
    \centering
    \includegraphics[width=0.99\linewidth]{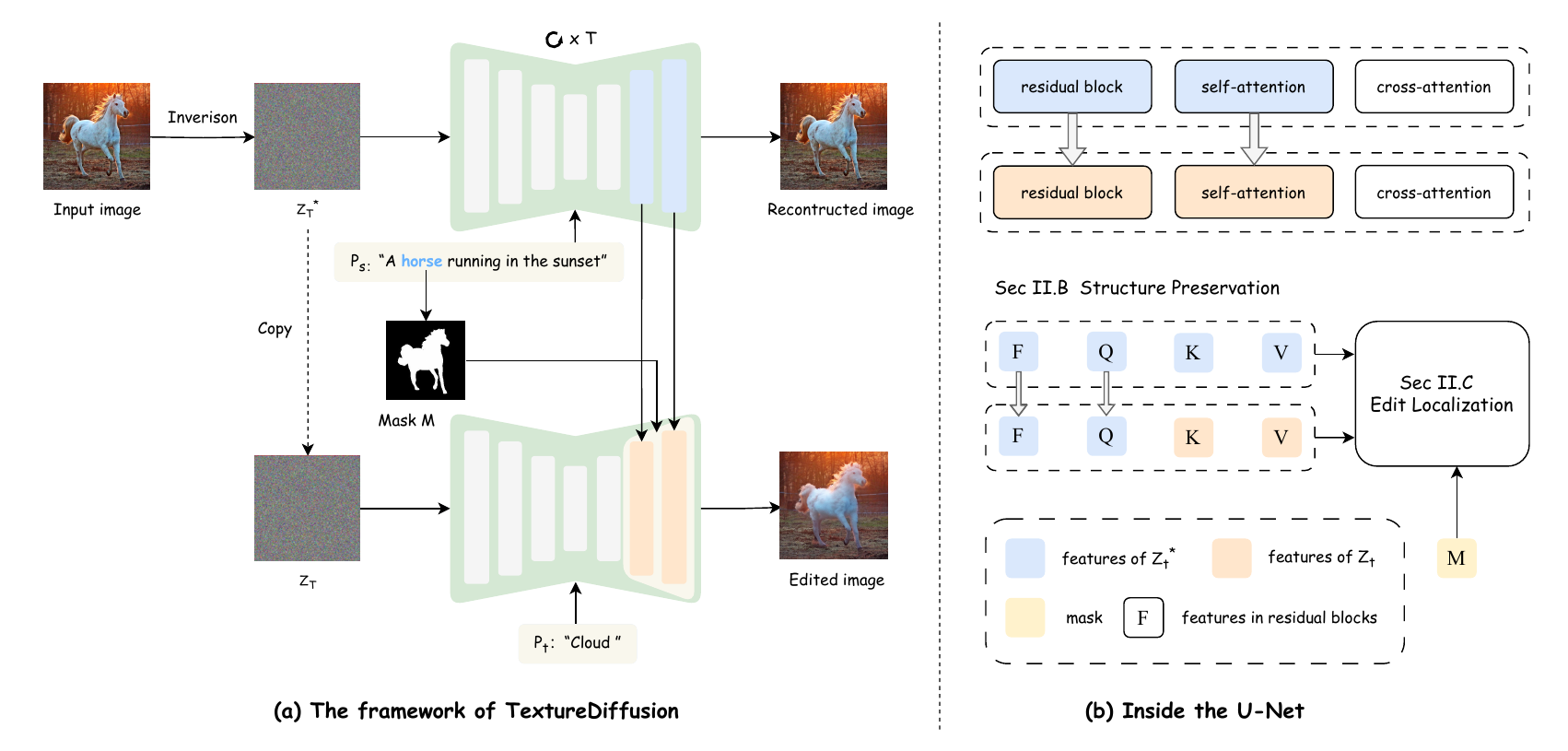}
    \caption{Pipeline of the proposed \textit{TextureDiffusion}. 
(a) Our method inverts the input image into an initial latent $Z_{T}^{*}$ and denoises it using DDIM sampling.
In the denoising process, we directly set the target prompt to \emph{``$<$texture$>$"}.
(b) For structure preservation, query features in self-attention and features in residual blocks are injected during the generation of the edited image. 
For edit localization, we utilize self-attention results and mask obtained from the cross-attention map.
}
    \vspace{-0.5cm}
    \label{fig:pipeline}
\end{figure*}

Thus our core idea is to directly set the target prompt to \emph{``$<$texture$>$"}, making the texture disentangled from the description of the input image. Based on this, we propose \textit{TextureDiffusion}, a tuning-free image editing method applied to various texture transfer. Initially, the target prompt is modified to make texture representation unrestricted. Subsequently, to preserve the structure of input image, query features in self-attention and features in residual blocks are injected during the generation of the edited image. Finally, to maintain the background, we introduce an edit localization technique which blends the self-attention results and the intermediate latents.

Our main contributions are summarized as follows. \textbf{1)} We propose a tuning-free image editing method named \textit{TextureDiffusion}, which is 
applied to various texture transfer. \textbf{2)} We directly set the target prompt to \emph{``$<$texture$>$"} to improve texture representation. \textbf{3)} Comprehensive experiments demonstrate that \textit{TextureDiffusion} can harmoniously transfer various textures with excellent structure and background preservation.

\begin{figure*}[!t]
    \centering
    \includegraphics[width=0.99\linewidth]{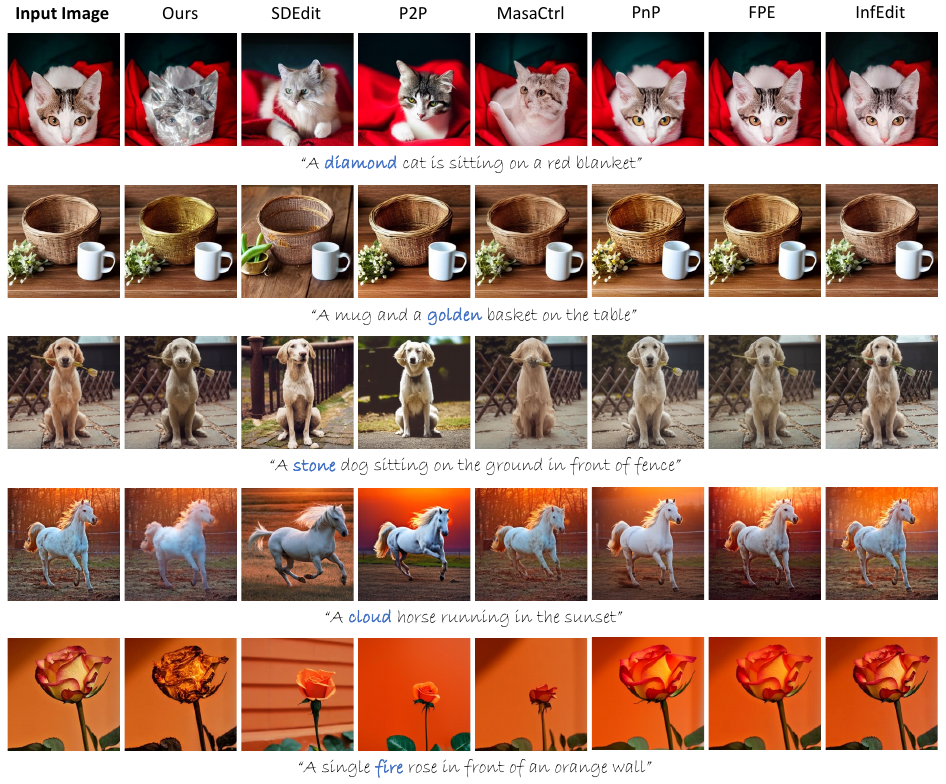}
    \caption{Results of qualitative comparisons. The blue word represents the texture. For our method, the target prompt is \emph{``$<$texture$>$"} only. For the other methods, the target prompt is a complete sentence. \textbf{Best viewed with zoom in.}}
    \vspace{-0.5cm}
    \label{fig:qualitative comparison}
\end{figure*}

\section{METHOD}
The pipeline of our method is depicted in Fig.~\ref{fig:pipeline}. Given an input image and a related text prompt $P_s$, our goal is to transfer various textures to the target object, aligned with the target text prompt $P_t$.
In this section, we first review the basic knowledge of diffusion models in Section~\ref{2A}.
Subsequently, a structure preservation module is introduced to maintain structural similarity between the edited and input image in Section~\ref{2B}.
Finally, we propose an edit localization technique to restrict the edit to the target object while keeping the rest unchanged in Section~\ref{2C}. 

\subsection{Preliminaries}\label{2A}
Diffusion models~\cite{ddpm,improved_ddpm, ddim, sohl2015deep, dhariwal2021diffusion} are generative models that can generate data by iterative denoising starting from Gaussian noise. It include a forward process and a reverse process. The forward process adds noise to the data sample $x_0$ at time step $t$ to generate the noisy sample $x_t$:
$q (x_t | x_0) = \mathcal {N} (x_t; \sqrt{\bar\alpha}_t x_0, (1-\bar\alpha_t)  I)$, 
where $\bar{\alpha}_t = \Pi_{i=1}^t \alpha_i$, $\alpha_i$ denotes the predefined noise schedule. The reverse process removes the noise from the previous sample $x_t$ to generate a clean sample $x_{t-1}$:
$p_\theta(x_{t-1}|x_t) = \mathcal{N}(x_{t-1}; \mu_{\theta}(x_t, t), \sigma_t),$
where $\sigma_t = \frac{1-\bar\alpha_{t-1}}{1-\bar\alpha_t} \beta_t$, $\beta_t=1-\alpha_t$, $\mu_\theta (x_t,t)= \frac{1}{\sqrt\alpha_t} (x_t - \frac{1-\alpha_t}{\sqrt{1-\bar\alpha_t}}\epsilon)$. Noise $\epsilon$ can be predicted by a neural network $\epsilon_\theta(x_t, t)$ trained on the objective: $L = E_{x_0, \epsilon, t}(\| \epsilon - \epsilon_\theta(x_t, t) \|).$
Additionally, when $\epsilon_\theta$ is conditioned on the text prompt $P$, it can be formulated as $\epsilon_\theta(x_t, t, P)$.
After doing so, the diffusion model can generate images that match the provided text prompt.

Our method is based on the state-of-the-art text-to-image model Stable Diffusion (SD)~\cite{latent_diffusion_model}. 
SD belongs to Latent Diffusion Models (LDMs) that performs the diffusion process in the latent space. SD is based on U-Net architecture~\cite{unet}.
The U-Net contains a series of basis blocks, each containing a residual block~\cite{resnet}, a self-attention module, and a cross-attention module~\cite{attention}.
Self-attention module contains important semantic information and its output can
be formulated as follows: 
\begin{equation}
    \text{Attention}(Q, K, V) = \text{Softmax}(\frac{QK^T}{\sqrt{d}})V,
\end{equation}
where $Q$, $K$, and $V$ are the query, key, and value features projected from spatial features with corresponding projection matrices.

\subsection{Structure Preservation}\label{2B}
After directly modifying the target prompt to \emph{``$<$texture$>$"}, information about the content of the input image is lost. Thus the structure of the input image needs to be preserved.

As mentioned in previous work~\cite{MasaCtrl, tune-a-video, patashnik2023localizing}, in the self-attention module of SD U-Net, the query features control the overall layout of the generated image, while the key and value features control the semantic contents. Therefore we inject the query features in the self-attention module into the generation process of the edited image and the result is shown in Fig.~\ref{fig:ablation study}. The structure of the input image is partially preserved after injecting the query features, but it is still insufficient and more structural information needs to be injected. Inspired by ~\cite{pnp}, which demonstrated that features in residual blocks contain the structural information of the input image, we further inject features in residual blocks and the experimental results are shown in Fig.~\ref{fig:ablation study}. The structure of the input image can be well maintained when query features in the self-attention module and features in residual blocks are injected at the same time.

In addition, since the generation process of the diffusion model is from the overall layout to the semantic details, structural information is injected only in the first and middle stages of the generation process. We do not inject the structural information in the later stages, which enables the texture details to be fully represented.

\begin{table*}[!t]
\caption{Quantitative results on the editing type of changing material on PIE-Bench.}
\centering
\normalsize
\begin{tabular}{@{}c|c|cccc|c@{}}
\toprule
\multirow{2}{*}[-0.1cm]{\textbf{Method}} & \textbf{Structure} & \multicolumn{4}{c|}{\textbf{Background Preservation}}             & \textbf{CLIP Similarity} \\ \cmidrule(l){2-7} 
                                 & \textbf{Distance}$_{10^3}$ $\downarrow$  & \textbf{PSNR} $\uparrow$  & \textbf{LPIPS}$_{10^3} \downarrow$ & \textbf{MSE}$_{10^4}\downarrow$   & \textbf{SSIM}$_{10^2}\uparrow$  & \textbf{Edited} $\uparrow$          \\ \midrule
\textbf{SDEdit}                  & 80.35              & 18.43          & 224.08         & 208.89         & 71.33          & 16.45                    \\
\textbf{P2P}                     & 72.89              & 18.52          & 183.54         & 187.98         & 75.76          & 15.47                    \\
\textbf{MasaCtrl}                & 28.53              & 23.55          & 87.61          & 67.3           & 84.45          & 15.92                    \\
\textbf{PnP}                     & 33.23              & 23.87          & 100.17         & 66.77          & 82.66          & 16.29                    \\
\textbf{FPE}                     & 11.57        & 26.79    & 55.93    & 37.29    & 87.23    & 15.73                    \\
\textbf{InfEdit}                 & 22.74              & 24.28          & 57.33          & 66.37          & 85.8           & 15.97                    \\ \midrule
\textbf{Ours}                    & \textbf{10.39}     & \textbf{31.22} & \textbf{31.99} & \textbf{14.92} & \textbf{90.08} & \textbf{16.88}              \\ \bottomrule
\end{tabular}
\vspace{-0.3cm}
\label{tab:quantitative results}
\end{table*}

\subsection{Edit Localization}\label{2C}
To localize the edit on the target object while keeping the rest unchanged, we introduce an
edit localization technique. 

Initially, the position of the target object must be identified.
Drawing inspiration from~\cite{p2p}, the cross-attention map contains location information of the prompt tokens.
Therefore, we aggregate cross-attention maps across all heads and layers of the spatial resolution of 16×16. Subsequently, we extract the map corresponding to the target object and binarize it to derive the mask $M$.

Since the self-attention module in SD U-Net contains important semantic information, we blend the self-attention results from the source image and the edited images:
\begin{align}
    \label{eq:mask_attn}
    R^l_s &= \text{Attention}(Q^l_s, K^l_s, V^l_s), \\
    R^l_t &= \text{Attention}(Q^l_s, K^l_t, V^l_t), \\
    \bar{R}^l &= R^l_s \odot M + R^l_{t} \odot (1 - M),
\end{align}
where $\odot$ represents the Hadamard product and $\bar{R}^l$ denotes the ultimate attention output.
To further keep the remainder unchanged, we blend the intermediate latents of the source and edited images:
\begin{align}
    \label{eq:mask_latent}
    Z_t = Z_t \odot M + Z_t^* \odot (1 - M),
\end{align}
where $Z_t$ denotes the intermediate latents of the edited image.
Using this edit localization technique, the edit is restricted to the target object, keeping the remainder unchanged.

\section{EXPERIMENTS}

We implement the proposed method on Stable Diffusion~\cite{latent_diffusion_model} using publicly available checkpoints v1.4.  During sampling, we apply DDIM~\cite{ddim} with 50 denoising steps and set a classifier-free guidance value of 7.5.
Query features insertion in self-attention module is performed in the first 40 steps and in layers 12 to 15 of U-Net.
Features insertion in residual blocks is performed in all steps and in layer 7 of U-Net.

\subsection{Comparisons with Previous Works}

We compare the proposed method to state-of-the-art baselines that can be applied to text-guided image editing tasks, including: SDEdit~\cite{sdedit}, P2P~\cite{p2p}, PnP~\cite{pnp}, MasaCtrl~\cite{MasaCtrl}, FPE~\cite{fpe}, and InfEdit~\cite{infedit}. We use their open-sourced codes to produce the editing results.

\textbf{Qualitative Experiments} \enspace As shown in Fig.~\ref{fig:qualitative comparison}, we present the qualitative results of our method compared with the baselines. SDEdit edits the input image by adding noise to it and then denoising it, but this process does not preserve the structure of the input image. P2P adds an additional cross-attention map corresponding to texture, which alters the structure of the input image and changes the shape of the target object. MasaCtrl applies mutual self-attention to preserve the contents of the input image, preventing changing the texture of the target object. PnP and FPE inject structural information from the input image to maintain the structure, and InfEdit uses virtual inversion to achieve efficient image reconstruction. However, among these methods, the description of the input image in the target prompt restricts the representation of the texture, preventing the texture to be successfully transferred. In contrast, our method successfully transfer various textures to the target object while keeping the remainder unchanged.

\textbf{Quantitative Experiments} \enspace The dataset is the editing type of changing material on PIE-Bench~\cite{pie-bench}. We find that some text prompts do not meet the standards for changing material, so we modify them.
To demonstrate the efficiency of our method, we employ six metrics including four aspects: structure distance~\cite{structure_distance}, background preservation (PSNR, LPIPS~\cite{LPIPS}, MSE, and SSIM~\cite{SSIM} outside the annotated editing mask), and edit prompt-image consistency (CLIP Similariy~\cite{CLIPSIM}) . Note that to evaluate whether the texture has been transferred to the target object, we set the prompt to \emph{``$<$texture$>$"} \ only and calculate the CLIP Similarity between the prompt and the target object region of edited image.

Tab.~\ref{tab:quantitative results} shows quantitative results of our method compared with the baselines. As seen, our method outperforms the baselines by achieving highest preservation of structure, highest preservation of background and highest fidelity to the prompt.

\begin{figure}[!t]
    \centering
    \includegraphics[width=\linewidth]{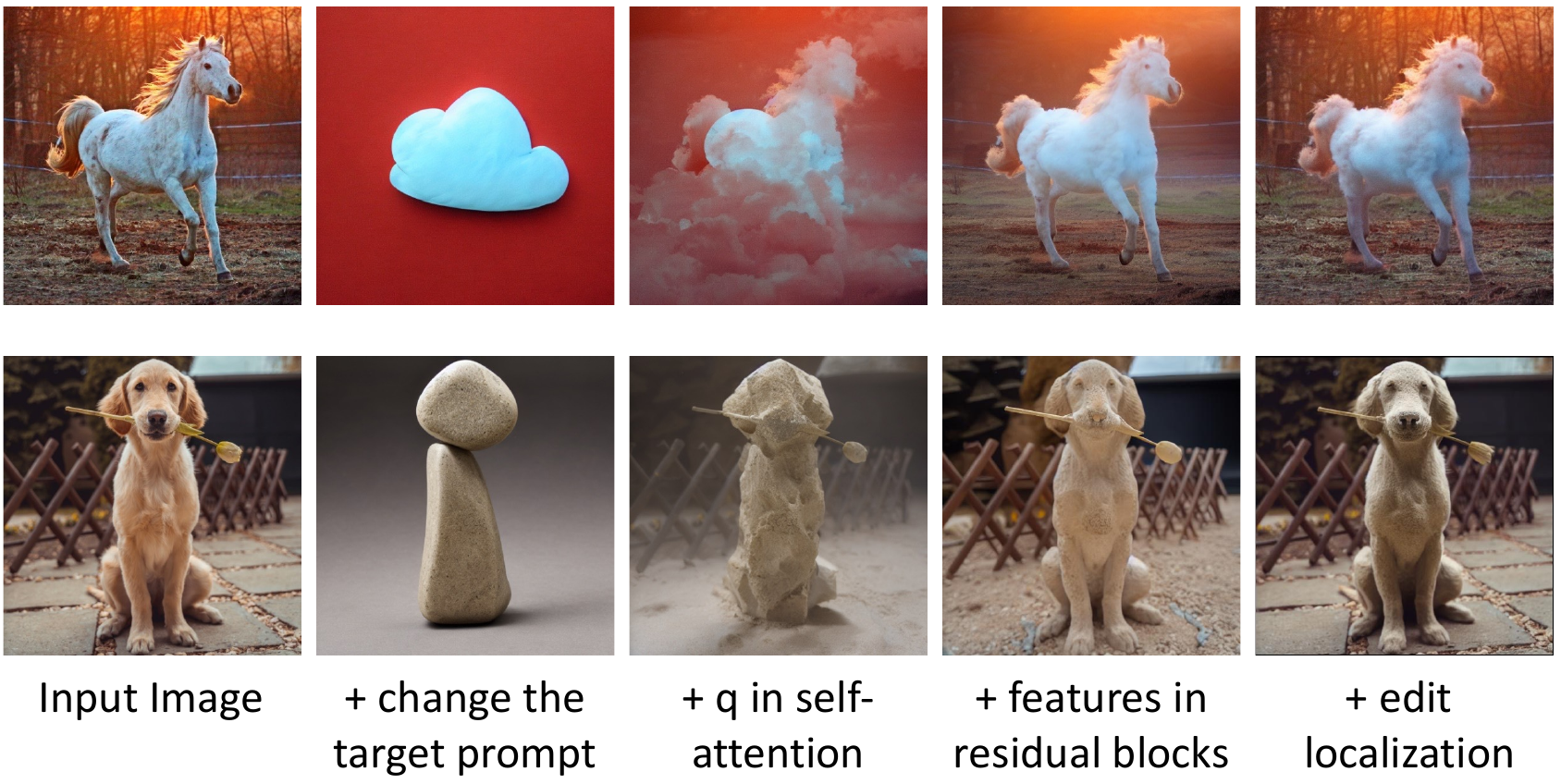}
    \caption{Results of ablation study.}
     \vspace{-0.5cm}
    \label{fig:ablation study}
\end{figure}

\subsection{Ablation Study}
We conduct an ablation study to validate the effectiveness of our designed core components and the results is shown in Fig.~\ref{fig:ablation study}. As seen, the texture can be fully represented when the target prompt is directly set to \emph{``$<$texture$>$"}. When both query features in self-attention module and features in residual blocks are added during the generation of the edited image, the structure of the input image is well preserved. When applying the proposed edit localization technique, the background is well retained.

\section{CONCLUSION}
We proposed \textit{TextureDiffusion}, a tuning-free image editing method applied to various texture transfer.
We enhanced the representation of complex textures by directly setting the target prompt to \emph{``$<$texture$>$"}. We also presented a structure preserve module and an edit localization technique.
Comprehensive experiments show that \textit{TextureDiffusion} can harmoniously transfer various textures with excellent structure background preservation.
Although we introduced the edit localization technique, the background is still slightly altered due to the upper limit of the image reconstruction quality of the variational autoencoder. We will explore transferring multiple textures simultaneously in the future.

\bibliographystyle{IEEEtran}
\bibliography{reference}

\begin{thebibliography}{10}
\providecommand{\url}[1]{#1}
\csname url@samestyle\endcsname
\providecommand{\newblock}{\relax}
\providecommand{\bibinfo}[2]{#2}
\providecommand{\BIBentrySTDinterwordspacing}{\spaceskip=0pt\relax}
\providecommand{\BIBentryALTinterwordstretchfactor}{4}
\providecommand{\BIBentryALTinterwordspacing}{\spaceskip=\fontdimen2\font plus
\BIBentryALTinterwordstretchfactor\fontdimen3\font minus \fontdimen4\font\relax}
\providecommand{\BIBforeignlanguage}[2]{{%
\expandafter\ifx\csname l@#1\endcsname\relax
\typeout{** WARNING: IEEEtran.bst: No hyphenation pattern has been}%
\typeout{** loaded for the language `#1'. Using the pattern for}%
\typeout{** the default language instead.}%
\else
\language=\csname l@#1\endcsname
\fi
#2}}
\providecommand{\BIBdecl}{\relax}
\BIBdecl

\bibitem{ramesh2021zero}
A.~Ramesh, M.~Pavlov, G.~Goh, S.~Gray, C.~Voss, A.~Radford, M.~Chen, and I.~Sutskever, ``Zero-shot text-to-image generation,'' in \emph{ICML}.\hskip 1em plus 0.5em minus 0.4em\relax PMLR, 2021, pp. 8821--8831.

\bibitem{nichol2021glide}
A.~Nichol, P.~Dhariwal, A.~Ramesh, P.~Shyam, P.~Mishkin, B.~McGrew, I.~Sutskever, and M.~Chen, ``Glide: Towards photorealistic image generation and editing with text-guided diffusion models,'' \emph{arXiv preprint arXiv:2112.10741}, 2021.

\bibitem{yu2022parti}
J.~Yu, Y.~Xu, J.~Y. Koh, T.~Luong, G.~Baid, Z.~Wang, V.~Vasudevan, A.~Ku, Y.~Yang, B.~K. Ayan \emph{et~al.}, ``Scaling autoregressive models for content-rich text-to-image generation,'' \emph{arXiv preprint arXiv:2206.10789}, 2022.

\bibitem{ramesh2022hierarchical}
A.~Ramesh, P.~Dhariwal, A.~Nichol, C.~Chu, and M.~Chen, ``Hierarchical text-conditional image generation with clip latents,'' \emph{arXiv preprint arXiv:2204.06125}, 2022.

\bibitem{balaji2022ediff}
Y.~Balaji, S.~Nah, X.~Huang, A.~Vahdat, J.~Song, K.~Kreis, M.~Aittala, T.~Aila, S.~Laine, B.~Catanzaro \emph{et~al.}, ``ediff-i: Text-to-image diffusion models with an ensemble of expert denoisers. corr, vol. abs/2211.01324 (2022),'' 2022.

\bibitem{chen2022distribution}
K.~Chen, J.~Song, S.~Liu, N.~Yu, Z.~Feng, G.~Han, and M.~Song, ``Distribution knowledge embedding for graph pooling,'' \emph{IEEE Transactions on Knowledge and Data Engineering}, 2022.

\bibitem{yang2023dynamic}
F.~Yang, S.~Yang, M.~A. Butt, J.~van~de Weijer \emph{et~al.}, ``Dynamic prompt learning: Addressing cross-attention leakage for text-based image editing,'' \emph{NeurIPS}, vol.~36, pp. 26\,291--26\,303, 2023.

\bibitem{Imagic}
B.~Kawar, S.~Zada, O.~Lang, O.~Tov, H.~Chang, T.~Dekel, I.~Mosseri, and M.~Irani, ``Imagic: Text-based real image editing with diffusion models,'' in \emph{CVPR}, 2023, pp. 6007--6017.

\bibitem{Instructpix2pix}
T.~Brooks, A.~Holynski, and A.~A. Efros, ``Instructpix2pix: Learning to follow image editing instructions,'' in \emph{CVPR}, 2023, pp. 18\,392--18\,402.

\bibitem{Text2live}
O.~Bar-Tal, D.~Ofri-Amar, R.~Fridman, Y.~Kasten, and T.~Dekel, ``Text2live: Text-driven layered image and video editing,'' in \emph{ECCV}.\hskip 1em plus 0.5em minus 0.4em\relax Springer, 2022.

\bibitem{diffusionclip}
G.~Kim, T.~Kwon, and J.~C. Ye, ``Diffusionclip: Text-guided diffusion models for robust image manipulation,'' in \emph{CVPR}, 2022, pp. 2426--2435.

\bibitem{multiedits}
M.~Huang, J.~Cai, S.~Jia, V.~S. Lokhande, and S.~Lyu, ``Multiedits: Simultaneous multi-aspect editing with text-to-image diffusion models,'' \emph{arXiv preprint arXiv:2406.00985}, 2024.

\bibitem{p2p}
A.~Hertz, R.~Mokady, J.~Tenenbaum, K.~Aberman, Y.~Pritch, and D.~Cohen-Or, ``Prompt-to-prompt image editing with cross attention control,'' \emph{arXiv preprint arXiv:2208.01626}, 2022.

\bibitem{pnp}
N.~Tumanyan, M.~Geyer, S.~Bagon, and T.~Dekel, ``Plug-and-play diffusion features for text-driven image-to-image translation,'' in \emph{CVPR}, 2023, pp. 1921--1930.

\bibitem{infedit}
S.~Xu, Y.~Huang, J.~Pan, Z.~Ma, and J.~Chai, ``Inversion-free image editing with natural language,'' \emph{arXiv preprint arXiv:2312.04965}, 2023.

\bibitem{ddpm}
J.~Ho, A.~Jain, and P.~Abbeel, ``Denoising diffusion probabilistic models,'' \emph{NeurIPS}, vol.~33, pp. 6840--6851, 2020.

\bibitem{improved_ddpm}
A.~Q. Nichol and P.~Dhariwal, ``Improved denoising diffusion probabilistic models,'' in \emph{ICML}.\hskip 1em plus 0.5em minus 0.4em\relax PMLR, 2021, pp. 8162--8171.

\bibitem{ddim}
J.~Song, C.~Meng, and S.~Ermon, ``Denoising diffusion implicit models,'' \emph{arXiv preprint arXiv:2010.02502}, 2020.

\bibitem{sohl2015deep}
J.~Sohl-Dickstein, E.~Weiss, N.~Maheswaranathan, and S.~Ganguli, ``Deep unsupervised learning using nonequilibrium thermodynamics,'' in \emph{ICML}.\hskip 1em plus 0.5em minus 0.4em\relax PMLR, 2015, pp. 2256--2265.

\bibitem{dhariwal2021diffusion}
P.~Dhariwal and A.~Nichol, ``Diffusion models beat gans on image synthesis,'' \emph{NeurIPS}, vol.~34, pp. 8780--8794, 2021.

\bibitem{latent_diffusion_model}
R.~Rombach, A.~Blattmann, D.~Lorenz, P.~Esser, and B.~Ommer, ``High-resolution image synthesis with latent diffusion models,'' in \emph{CVPR}, 2022, pp. 10\,684--10\,695.

\bibitem{unet}
O.~Ronneberger, P.~Fischer, and T.~Brox, ``U-net: Convolutional networks for biomedical image segmentation,'' in \emph{MICCAI}.\hskip 1em plus 0.5em minus 0.4em\relax Springer, 2015, pp. 234--241.

\bibitem{resnet}
K.~He, X.~Zhang, S.~Ren, and J.~Sun, ``Deep residual learning for image recognition,'' in \emph{CVPR}, 2016, pp. 770--778.

\bibitem{attention}
A.~Vaswani, N.~Shazeer, N.~Parmar, J.~Uszkoreit, L.~Jones, A.~N. Gomez, {\L}.~Kaiser, and I.~Polosukhin, ``Attention is all you need,'' \emph{NeurIPS}, vol.~30, 2017.

\bibitem{MasaCtrl}
M.~Cao, X.~Wang, Z.~Qi, Y.~Shan, X.~Qie, and Y.~Zheng, ``Masactrl: Tuning-free mutual self-attention control for consistent image synthesis and editing,'' in \emph{ICCV}, October 2023, pp. 22\,560--22\,570.

\bibitem{tune-a-video}
J.~Z. Wu, Y.~Ge, X.~Wang, S.~W. Lei, Y.~Gu, Y.~Shi, W.~Hsu, Y.~Shan, X.~Qie, and M.~Z. Shou, ``Tune-a-video: One-shot tuning of image diffusion models for text-to-video generation,'' in \emph{ICCV}, 2023, pp. 7623--7633.

\bibitem{patashnik2023localizing}
O.~Patashnik, D.~Garibi, I.~Azuri, H.~Averbuch-Elor, and D.~Cohen-Or, ``Localizing object-level shape variations with text-to-image diffusion models,'' in \emph{ICCV}, 2023, pp. 23\,051--23\,061.

\bibitem{sdedit}
C.~Meng, Y.~He, Y.~Song, J.~Song, J.~Wu, J.-Y. Zhu, and S.~Ermon, ``Sdedit: Guided image synthesis and editing with stochastic differential equations,'' in \emph{ICLR}, 2022.

\bibitem{fpe}
B.~Liu, C.~Wang, T.~Cao, K.~Jia, and J.~Huang, ``Towards understanding cross and self-attention in stable diffusion for text-guided image editing,'' in \emph{CVPR}, 2024, pp. 7817--7826.

\bibitem{pie-bench}
X.~Ju, A.~Zeng, Y.~Bian, S.~Liu, and Q.~Xu, ``Pnp inversion: Boosting diffusion-based editing with 3 lines of code,'' in \emph{ICLR}, 2024.

\bibitem{structure_distance}
N.~Tumanyan, O.~Bar-Tal, S.~Bagon, and T.~Dekel, ``Splicing vit features for semantic appearance transfer,'' in \emph{CVPR}, 2022, pp. 10\,748--10\,757.

\bibitem{LPIPS}
R.~Zhang, P.~Isola, A.~A. Efros, E.~Shechtman, and O.~Wang, ``The unreasonable effectiveness of deep features as a perceptual metric,'' in \emph{CVPR}, 2018, pp. 586--595.

\bibitem{SSIM}
Z.~Wang, A.~C. Bovik, H.~R. Sheikh, and E.~P. Simoncelli, ``Image quality assessment: from error visibility to structural similarity,'' \emph{IEEE transactions on image processing}, vol.~13, no.~4, pp. 600--612, 2004.

\bibitem{CLIPSIM}
C.~Wu, L.~Huang, Q.~Zhang, B.~Li, L.~Ji, F.~Yang, G.~Sapiro, and N.~Duan, ``Godiva: Generating open-domain videos from natural descriptions,'' \emph{arXiv preprint arXiv:2104.14806}, 2021.

\end{thebibliography}

\end{document}